\documentclass[runningheads,a4paper]{llncs}

\usepackage{amssymb}
\usepackage{graphicx}
\usepackage{placeins}
\usepackage{color}
\usepackage{fixltx2e}
\usepackage{url}
\usepackage{placeins}
\usepackage[table]{xcolor}
\usepackage{multirow}
\usepackage[hidelinks]{hyperref}
\usepackage{todonotes}
\usepackage[export]{adjustbox}
\definecolor{red}{rgb}{0.79509804490953684,0.20098039414733648,0.20686274673789687}
\definecolor{blue}{rgb}{0.27900615537575646,0.48814110222692564,0.65689543960433383}
\definecolor{green}{rgb}{0.35022877099759442,0.63449636544374854,0.34172819480299954}
\definecolor{purple}{rgb}{0.56810267475598009,0.34858709455412984,0.591251461777617}
\definecolor{orange}{rgb}{0.87539215686858873,0.50482698971149964,0.12774509808012069}
\definecolor{yellow}{rgb}{0.8939061906626995,0.88929258034948033,0.29840446252594988}
\definecolor{pink}{rgb}{0.90225298544939825,0.56472128699807567,0.74206845059114346}
\urldef{\mail}\path|{p.moeskops, j.m.wolterink, m.viergever, i.isgum}@umcutrecht.nl|

\newcommand{\keywords}[1]{\par\addvspace\baselineskip
\noindent\keywordname\enspace\ignorespaces#1}

\newcommand{\repeatthanks}{\textsuperscript{\thefootnote}}

\begin{document}

\mainmatter  

\title{Deep Learning for Multi-Task Medical Image Segmentation in Multiple Modalities}

\titlerunning{Deep Learning for Multi-Task Medical Image Segmentation}
%
%
\author{Pim Moeskops\inst{1,2}\thanks{Both authors contributed equally.} \and Jelmer M. Wolterink\inst{1}\repeatthanks \and Bas H.M. van der Velden\inst{1} \and \\ Kenneth G.A. Gilhuijs\inst{1} \and Tim Leiner\inst{3} \and Max A. Viergever\inst{1} \and Ivana I\v{s}gum\inst{1}}

\authorrunning{P. Moeskops, J.M. Wolterink et al.}


\institute{Image Sciences Institute, University Medical Center Utrecht, The Netherlands \and Medical Image Analysis, Eindhoven University of Technology, The Netherlands \and Department of Radiology, University Medical Center Utrecht, The Netherlands}


%
%

\maketitle

\begin{abstract}
Automatic segmentation of medical images is an important task for many clinical applications. In practice, a wide range of anatomical structures are visualised using different imaging modalities. In this paper, we investigate whether a single convolutional neural network (CNN) can be trained to perform different segmentation tasks. 

A single CNN is trained to segment six tissues in MR brain images, the pectoral muscle in MR breast images, and the coronary arteries in cardiac CTA. The CNN therefore learns to identify the imaging modality, the visualised anatomical structures, and the tissue classes. 

For each of the three tasks (brain MRI, breast MRI and cardiac CTA), this combined training procedure resulted in a segmentation performance equivalent to that of a CNN trained specifically for that task, demonstrating the high capacity of CNN architectures. Hence, a single system could be used in clinical practice to automatically perform diverse segmentation tasks without task-specific training.\let\thefootnote\relax\footnote{This paper has been published in October 2016 as: Moeskops, P., Wolterink, J.M., van der Velden, B.H.M., Gilhuijs, K.G.A., Leiner, T., Viergever, M.A., and I\v{s}gum, I. (2016). Deep learning for multi-task medical image segmentation in multiple modalities. In: \textit{Medical Image Computing and Computer-Assisted Intervention - MICCAI 2016,} Part II, LNCS 9901, pp. 478-–486}

\keywords{Deep learning, Convolutional neural networks, Medical image segmentation, Brain MRI, Breast MRI, Cardiac CTA}
\end{abstract}

\section{Introduction}
Automatic segmentation is an important task in medical images acquired with different modalities visualising a wide range of anatomical structures.
A common approach to automatic segmentation is the use of supervised voxel classification, where a classifier is trained to assign a class label to each voxel. The classical approach to supervised classification is to train a classifier that discriminates between tissue classes based on a set of hand-crafted features. In contrast to this approach, convolutional neural networks (CNNs) automatically extract features that are optimised for the classification task at hand. 
CNNs have been successfully applied to medical image segmentation of e.g. knee cartilage \cite{Pras13}, brain regions \cite{Breb15,Moes16}, the pancreas \cite{Roth15a}, and coronary artery calcifications \cite{Wolt15}. Each of these studies employed CNNs, but problem-specific optimisations with respect to the network architecture were still performed and networks were only trained to perform one specific task.


CNNs have not only been used for processing of medical images, but also for natural images. CNN architectures designed for image classification in natural images \cite{Kriz12} have shown great generalisability for divergent tasks such as image segmentation \cite{Shel16}, object detection \cite{Girs16}, and object localisation in medical image analysis \cite{Vos16}. Hence, CNN architectures may have the flexibility to be used for different tasks with limited modifications. 

In this study, we first investigate the feasibility of using a single CNN \textit{architecture} for different medical image segmentation tasks in different imaging modalities visualising different anatomical structures. Secondly, we investigate the feasibility of using a single \textit{trained instance} of this CNN architecture for different segmentation tasks.
Such a system would be able to perform multiple tasks in different modalities without problem-specific tailoring of the network architecture or hyperparameters. 
Hence, the network recognises the modality of the image, the anatomy visualised in the image, and the tissues of interest. 
We demonstrate this concept using three different and potentially adversarial medical image segmentation problems: segmentation of six brain tissues in brain MRI, pectoral muscle segmentation in breast MRI, and coronary artery segmentation in cardiac CT angiography (CTA). 

\begin{figure}[t]
\includegraphics[width=\textwidth]{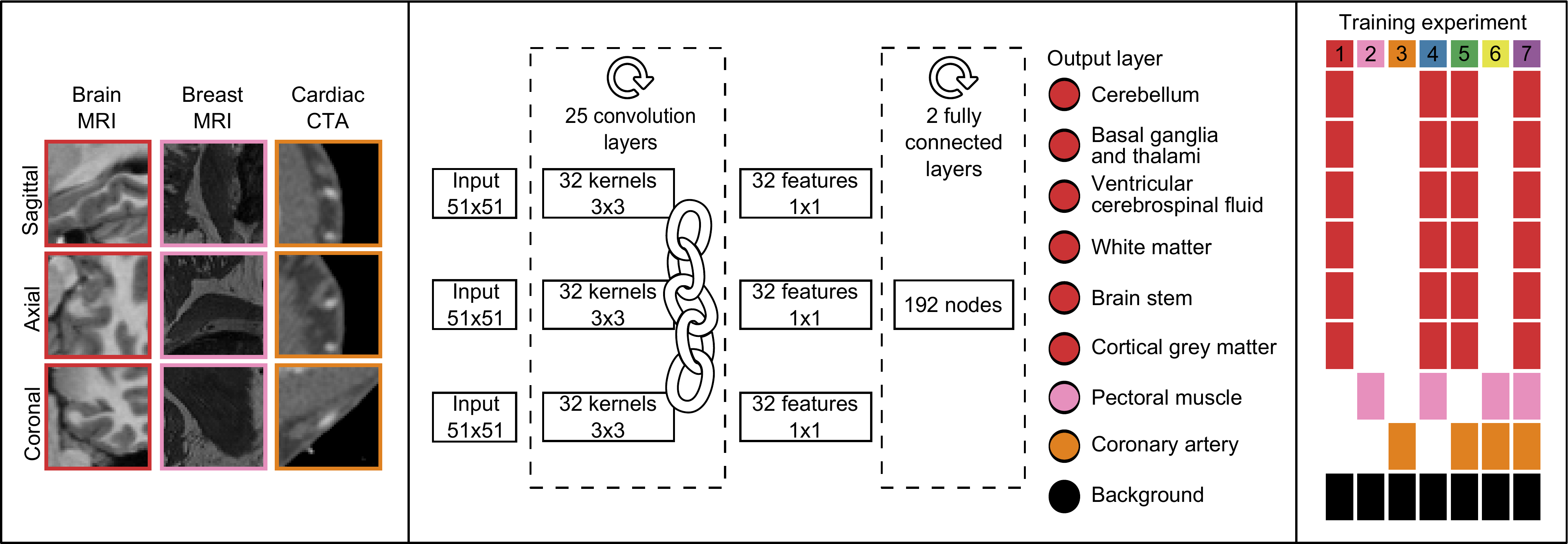}
\caption{Example 51$\times$51 triplanar input patches (\textit{left}). CNN architecture with 25 shared convolution layers, 2 fully connected layers and an output layer with at most 9 classes, including a background class common among tasks (\textit{centre}). Output classes included in each training experiment (\textit{right}).}
\label{fig:architecture}
\end{figure}

\begin{figure}[th]
\centering
\includegraphics[trim=5mm 0mm 15mm 5mm,clip,width=0.49\textwidth]{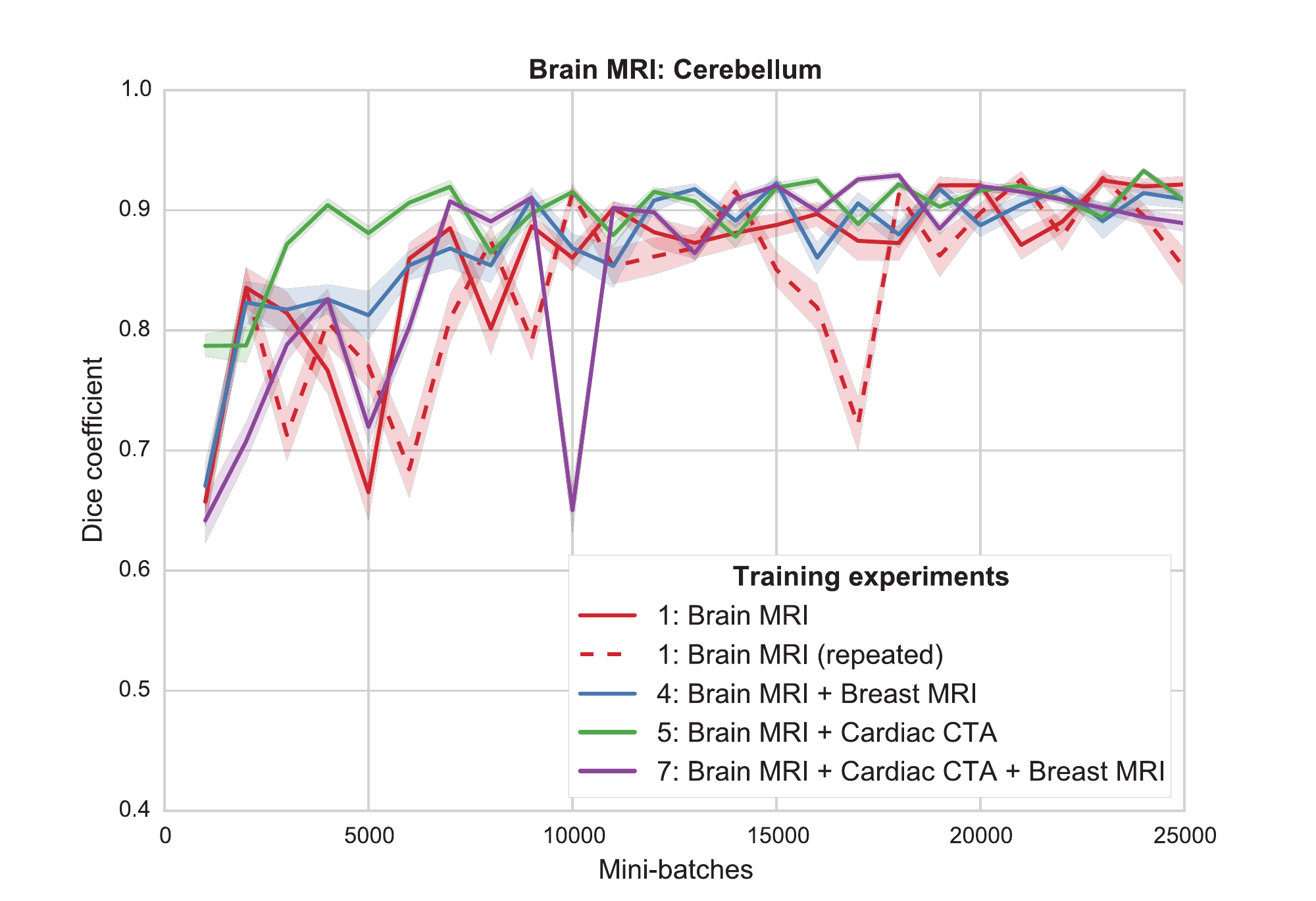} 
\includegraphics[trim=5mm 0mm 15mm 5mm,clip,width=0.49\textwidth]{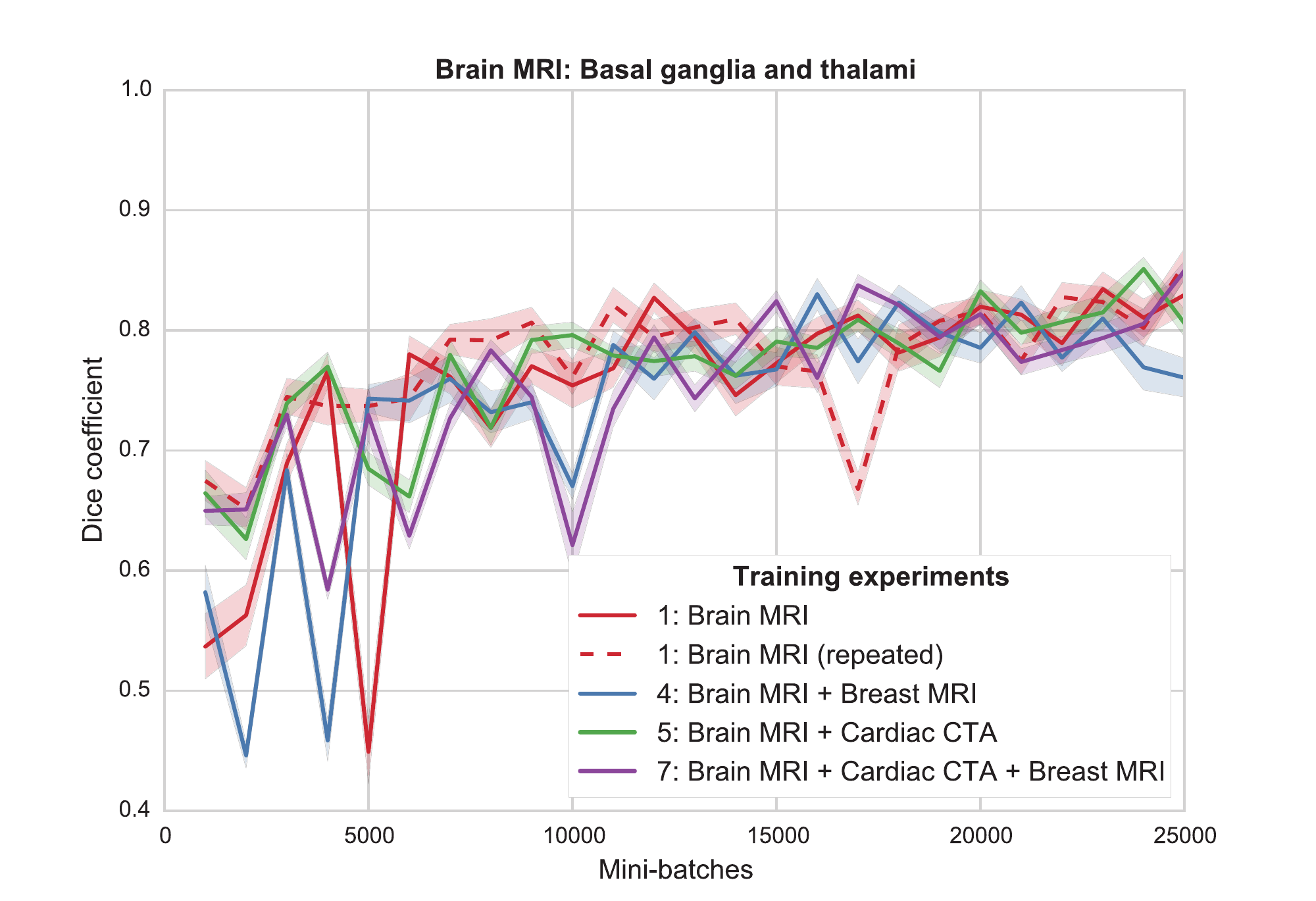}\\ 
\includegraphics[trim=5mm 0mm 15mm 5mm,clip,width=0.49\textwidth]{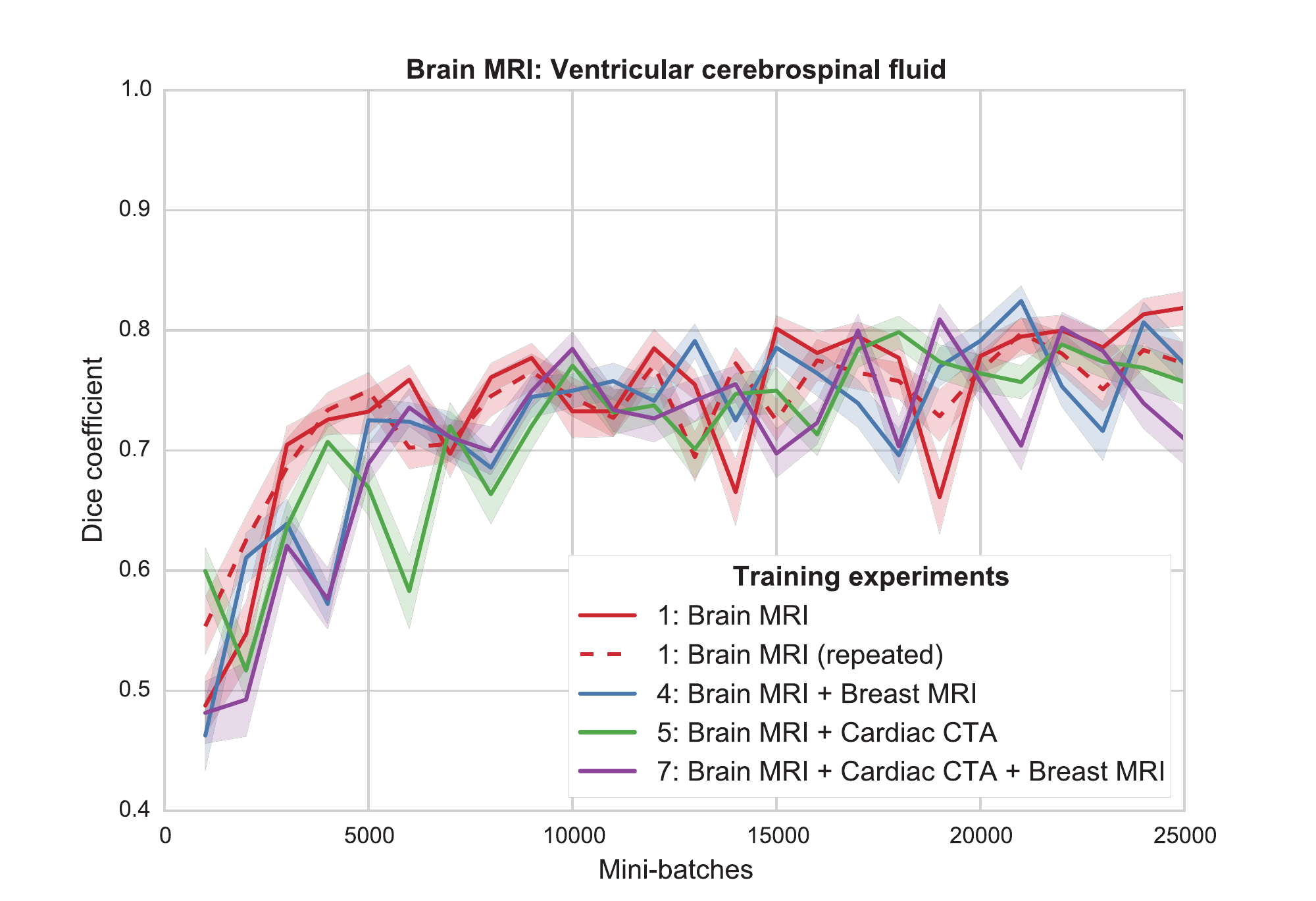} 
\includegraphics[trim=5mm 0mm 15mm 5mm,clip,width=0.49\textwidth]{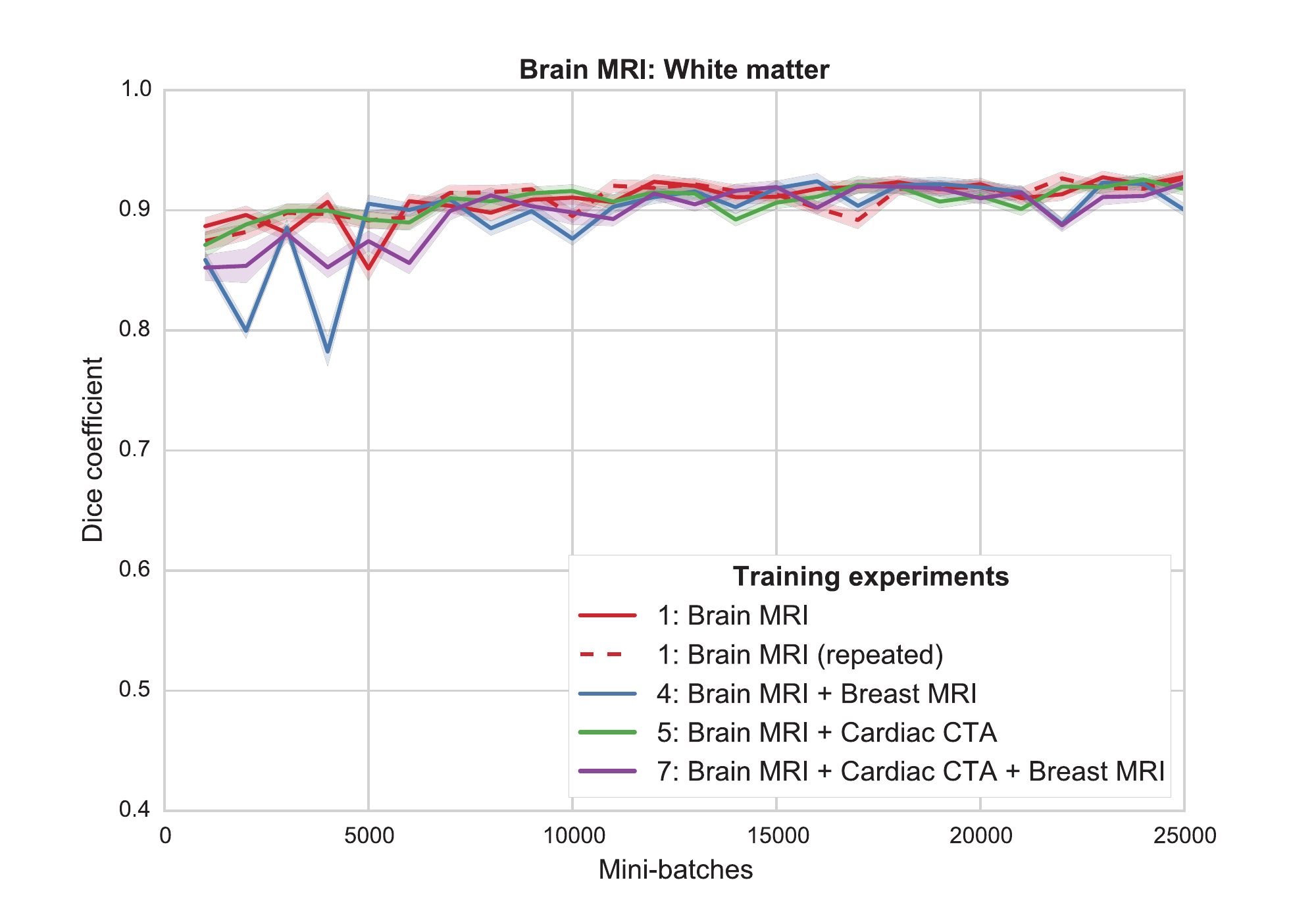}\\ 
\includegraphics[trim=5mm 0mm 15mm 5mm,clip,width=0.49\textwidth]{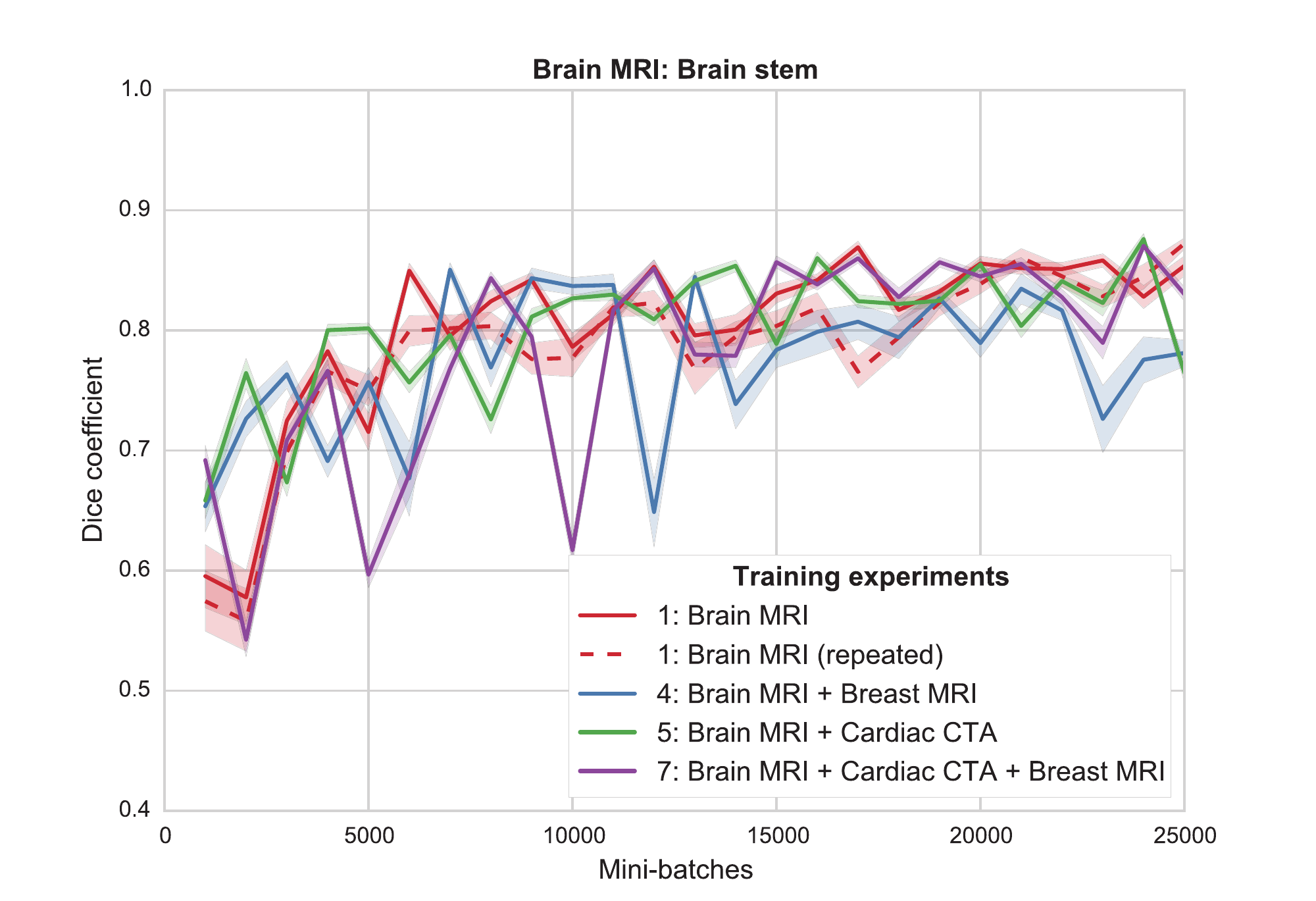} 
\includegraphics[trim=5mm 0mm 15mm 5mm,clip,width=0.49\textwidth]{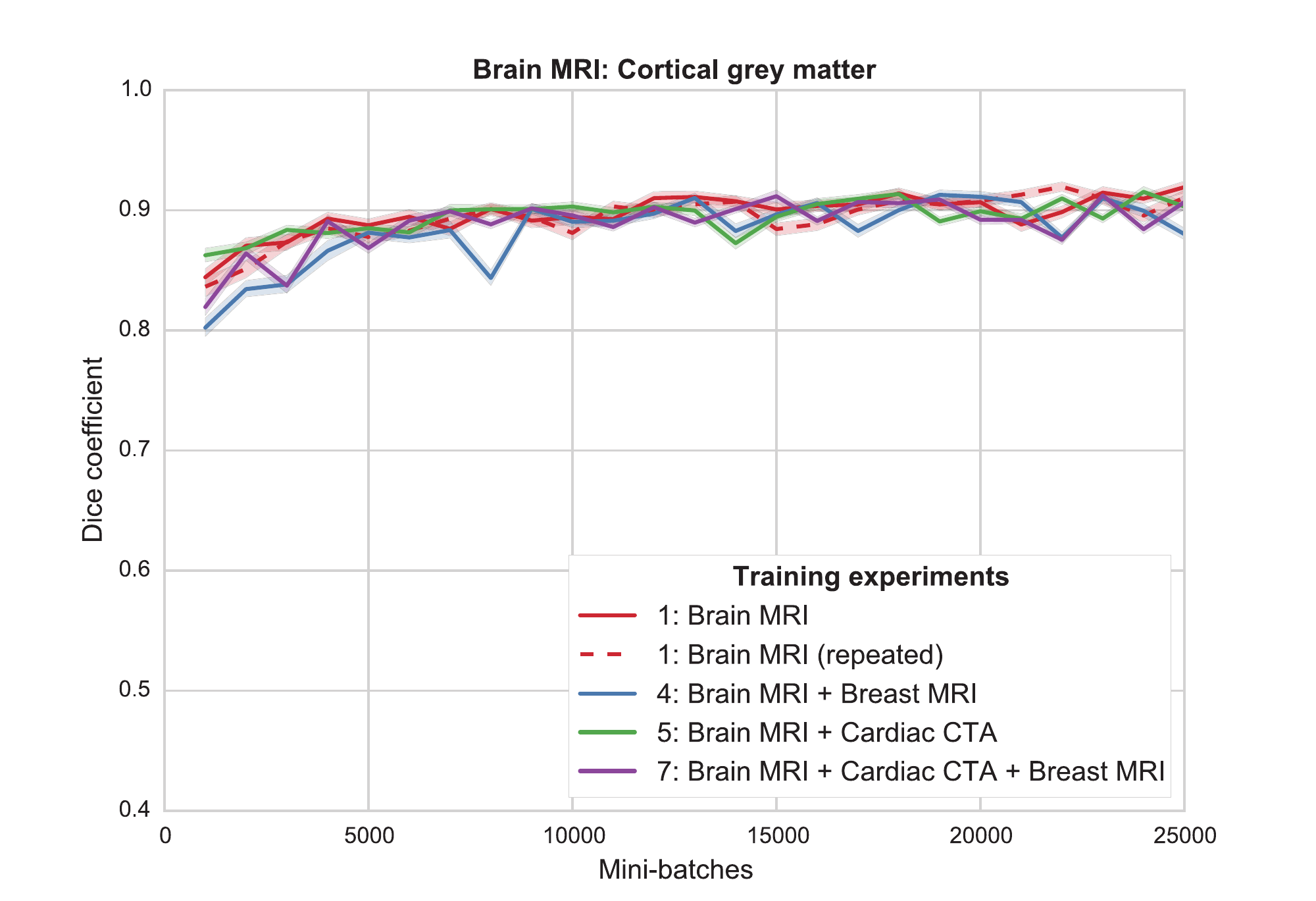} \\
\includegraphics[trim=5mm 0mm 15mm 5mm,clip,width=0.49\textwidth]{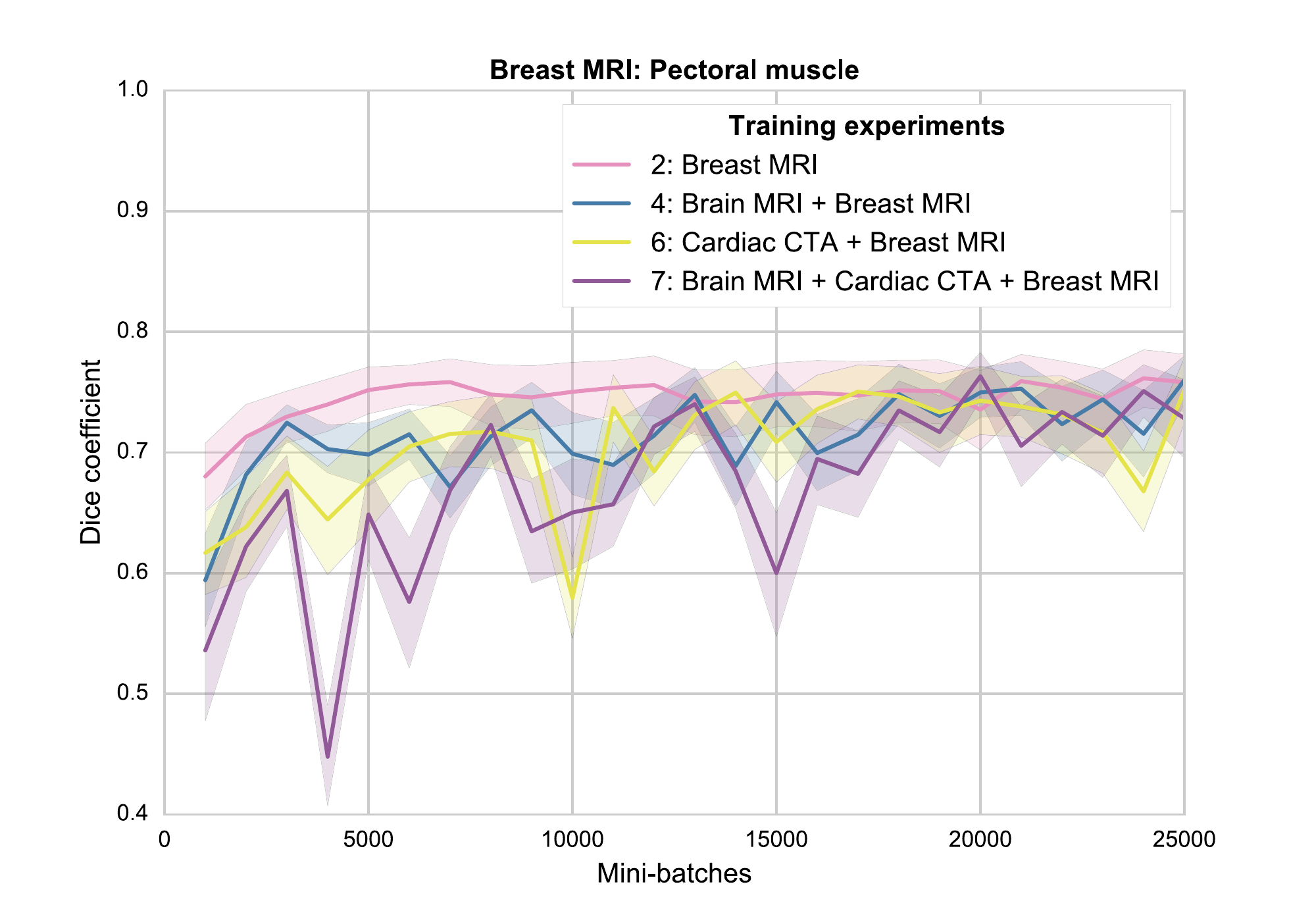} 
\includegraphics[trim=5mm 0mm 15mm 5mm,clip,width=0.49\textwidth]{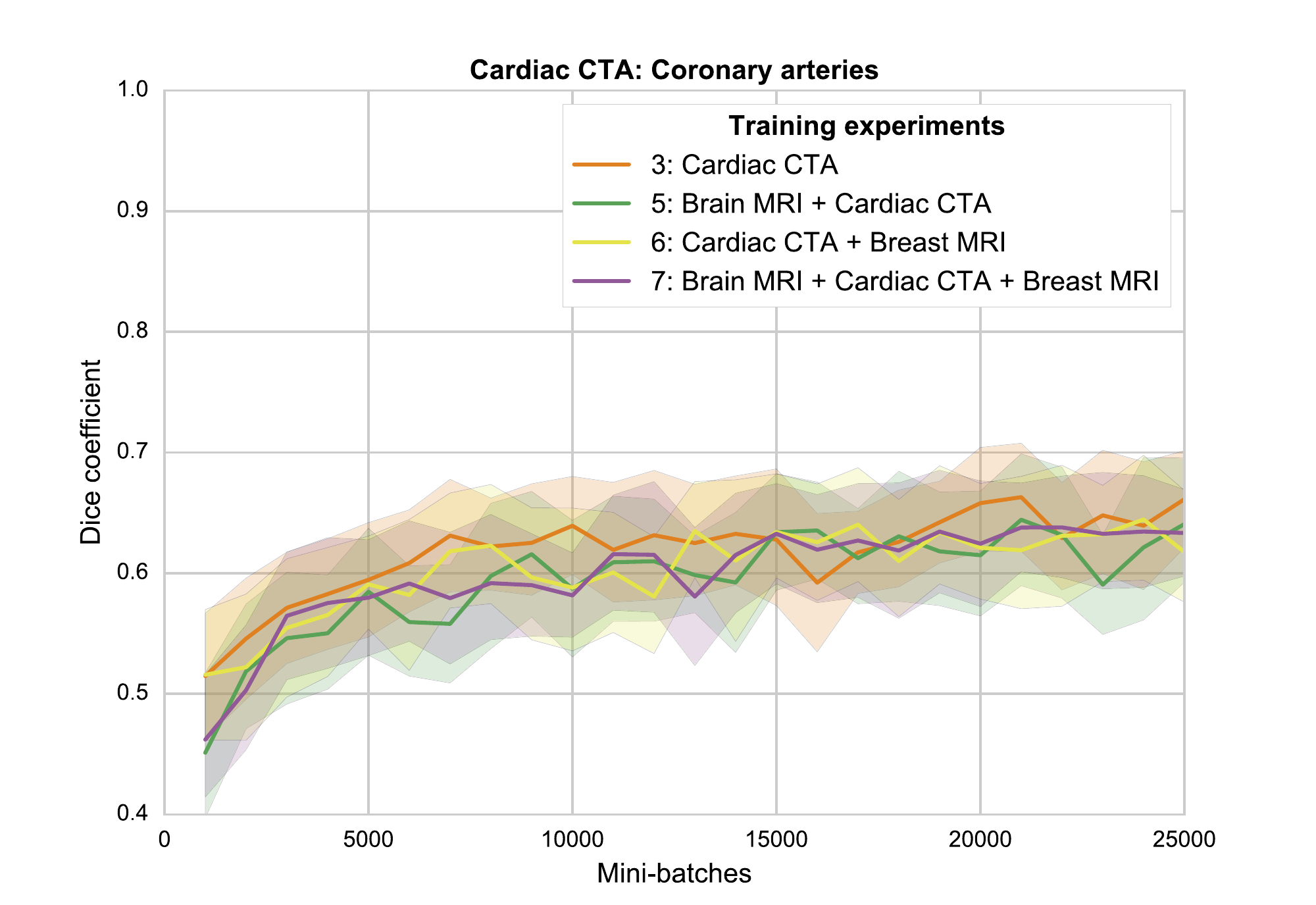} 
\caption{Learning curves showing Dice coefficients for tissue segmentation in brain MRI (\textit{top three rows}), breast MRI (\textit{bottom left}), and cardiac CTA (\textit{bottom right}), reported at 1000 mini-batch intervals for experiments including that task. The line colours correspond to the training experiments in Fig. \ref{fig:architecture}.} 
\label{fig:evaluation}
\end{figure}

\section{Data}

\paragraph{Brain MRI --}
34 T\textsubscript{1}-weighted MR brain images from the OASIS project \cite{Marc07} were acquired on a Siemens Vision 1.5 T scanner, as provided by the MICCAI challenge on multi-atlas labelling \cite{Land12}\footnote{\url{https://masi.vuse.vanderbilt.edu/workshop2012}}. The images were acquired with voxel sizes of 1.0$\times$1.0$\times$1.25 mm\textsuperscript{3} and resampled to isotropic voxel sizes of 1.0$\times$1.0$\times$1.0 mm\textsuperscript{3}. 
The images were manually segmented, in the coronal plane, into 134 classes that were, for the purpose of this paper, combined into six commonly used tissue classes: white matter, cortical grey matter, basal ganglia and thalami, ventricular cerebrospinal fluid, cerebellum, and brain stem.

\paragraph{Breast MRI --}34 T\textsubscript{1}-weighted MR breast images were acquired on a Siemens Magnetom 1.5 T scanner with a dedicated double breast array coil \cite{Veld15}. The images were acquired with in-plane voxel sizes between 1.21 and 1.35 mm and slice thicknesses between 1.35 and 1.69 mm. All images were resampled to isotropic voxel sizes corresponding to their in-plane voxel size.
The pectoral muscle was manually segmented in the axial plane by contour drawing. 

\paragraph{Cardiac CTA --}
Ten cardiac CTA scans were acquired on a 256-detector row Philips Brilliance iCT scanner using 120 kVp and 200-300 mAs, with ECG-triggering and contrast enhancement.
The reconstructed images had between 0.4 and 0.5 mm in-plane voxel sizes and 0.45/0.90 mm slice spacing/thickness. All images were resampled to isotropic 0.45$\times$0.45$\times$0.45 mm\textsuperscript{3} voxel size.
To set a manual reference standard, a human observer traversed the scan in the craniocaudal direction and painted voxels in the main coronary arteries and their branches in the axial plane.

\section{Method}
All voxels in the images were labelled by a CNN using seven different training experiments (Fig. \ref{fig:architecture}). 

\subsection{CNN Architecture}
For each voxel, three orthogonal (axial, sagittal, and coronal) patches of 51$\times$51 voxels centred at the target voxel were extracted. For each of these three patches, features were determined using a deep stack of convolution layers. Each convolution layer contained 32 small (3$\times$3 voxels) convolution kernels for a total of 25 convolution layers \cite{Simo14}. To prevent over- or undersegmentation of structures due to translational invariance, no subsampling layers were used. To reduce the number of trainable parameters in the network and hence the risk of over-fitting, the same stack of convolutional layers was used for the axial, sagittal and coronal patches.

The output of the convolution layers were 32 features for each of the three orthogonal input patches, hence, 96 features in total. These features were input to two subsequent fully connected layers, each with 192 nodes. The second fully connected layer was connected to a softmax classification layer. Depending on the tasks of the network, this layer contained 2, 3, 7, 8 or 9 output nodes. 
The fully connected layers were implemented as 1$\times$1 voxel convolutions, to allow fast processing of arbitrarily sized images.
Exponential linear units \cite{Clev16} were used for all non-linear activation functions. Batch normalisation \cite{Ioff15} was used on all layers and dropout \cite{Sriv14} was used on the fully connected layers. 

\begin{figure}[!th]
\centering
\fcolorbox{black}{black}{\includegraphics[trim=5mm 15mm 25mm 10mm,clip,width=0.27\textwidth]{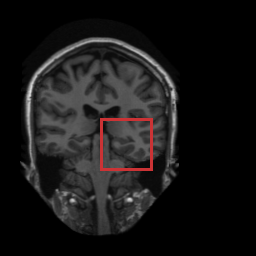}}
\fcolorbox{black}{black}{\includegraphics[trim=5mm 15mm 25mm 10mm,clip,width=0.27\textwidth]{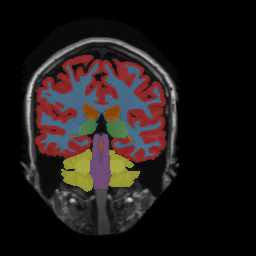}}
\fcolorbox{red}{black}{\includegraphics[trim=5mm 15mm 25mm 10mm,clip,width=0.27\textwidth]{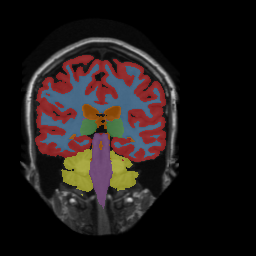}}\\
\fcolorbox{blue}{black}{\includegraphics[trim=5mm 15mm 25mm 10mm,clip,width=0.27\textwidth]{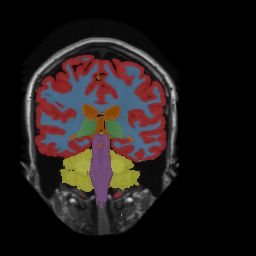}}
\fcolorbox{green}{black}{\includegraphics[trim=5mm 15mm 25mm 10mm,clip,width=0.27\textwidth]{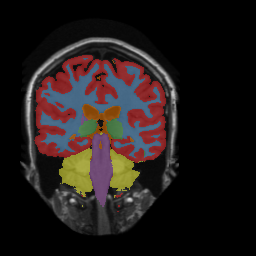}}
\fcolorbox{purple}{black}{\includegraphics[trim=5mm 15mm 25mm 10mm,clip,width=0.27\textwidth]{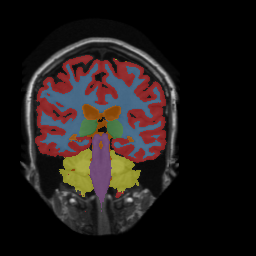}}\\
\fcolorbox{black}{black}{\includegraphics[trim=12mm 0mm 15mm 0mm,clip,width=0.27\textwidth]{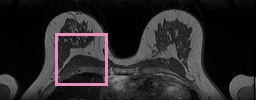}}
\fcolorbox{black}{black}{\includegraphics[trim=12mm 0mm 15mm 0mm,clip,width=0.27\textwidth]{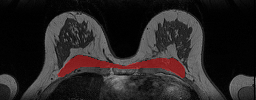}}
\fcolorbox{pink}{black}{\includegraphics[trim=12mm 0mm 15mm 0mm,clip,width=0.27\textwidth]{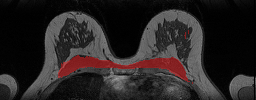}}\\
\fcolorbox{yellow}{black}{\includegraphics[trim=12mm 0mm 15mm 0mm,clip,width=0.27\textwidth]{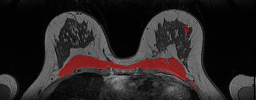}}
\fcolorbox{blue}{black}{\includegraphics[trim=12mm 0mm 15mm 0mm,clip,width=0.27\textwidth]{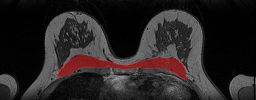}}
\fcolorbox{purple}{black}{\includegraphics[trim=12mm 0mm 15mm 0mm,clip,width=0.27\textwidth]{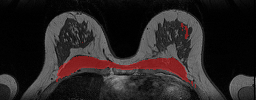}}\\
\fcolorbox{black}{black}{\includegraphics[trim=30mm 40mm 30mm 20mm,clip,width=0.27\textwidth]{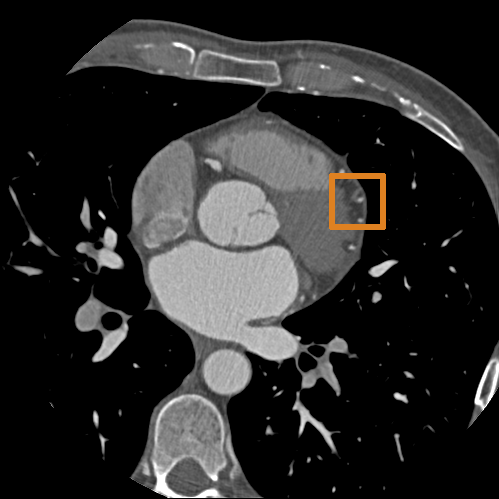}}
\fcolorbox{black}{black}{\includegraphics[trim=30mm 40mm 30mm 20mm,clip,width=0.27\textwidth]{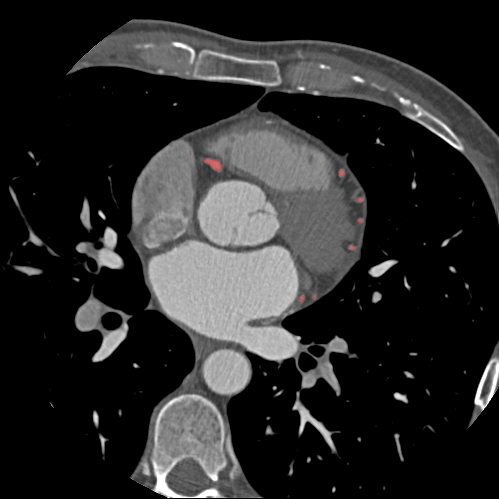}}
\fcolorbox{orange}{black}{\includegraphics[trim=30mm 40mm 30mm 20mm,clip,width=0.27\textwidth]{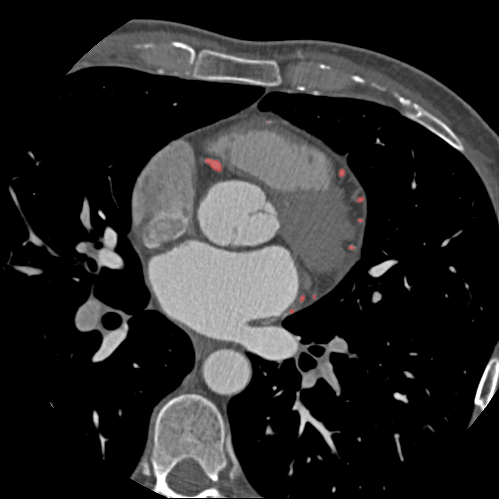}}\\
\fcolorbox{yellow}{black}{\includegraphics[trim=30mm 40mm 30mm 20mm,clip,width=0.27\textwidth]{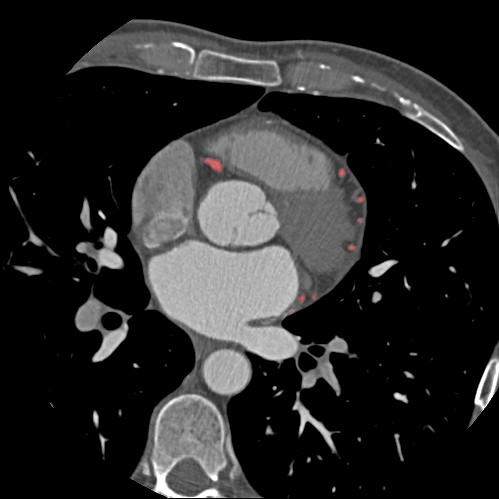}}
\fcolorbox{green}{black}{\includegraphics[trim=30mm 40mm 30mm 20mm,clip,width=0.27\textwidth]{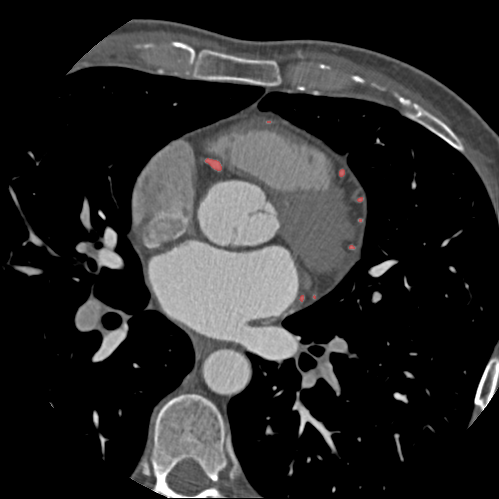}}
\fcolorbox{purple}{black}{\includegraphics[trim=30mm 40mm 30mm 20mm,clip,width=0.27\textwidth]{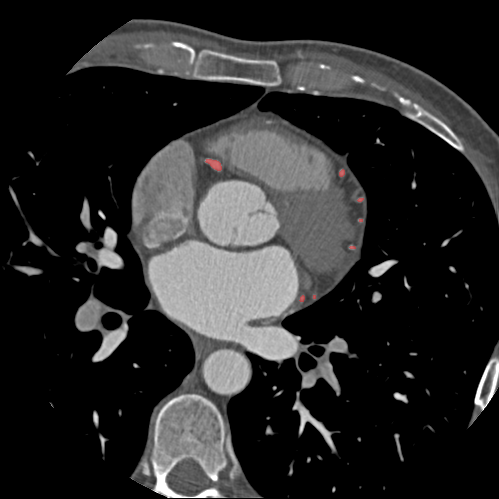}}\\
\caption{Example segmentations for (\textit{top to bottom}) brain MRI, breast MRI, and cardiac CTA. Shown for each task: (\textit{left to right, first row}) image with an input patch as shown in Fig. \ref{fig:architecture}, reference standard, segmentation by task-specific training, (\textit{left to right, second row}) two segmentations by networks with an additional task, segmentation by a network combining all tasks. The coloured borders correspond to the training experiments in Fig. \ref{fig:architecture} and Fig. \ref{fig:evaluation}.} 
\label{fig:segmentations}
\end{figure}


\subsection{Training Experiments}
The same model was trained for each combination of the three tasks. In total seven training experiments were performed (Fig. \ref{fig:architecture}, right): three networks were trained to perform one task (Experiments 1--3), three networks were trained to perform two tasks (Experiments 4--6), and one network was trained to perform three tasks (Experiment 7). The number of output nodes in the CNN was modified accordingly. In each experiment, background classes of the target tasks were merged into one class.

Each CNN was trained using mini-batch learning. A mini-batch contained 210 samples, equally balanced over the tasks of the network. For each task, the training samples were randomly drawn from all training images, balanced over the task-specific classes. All voxels with image intensity $>$ 0 were considered samples.
The network parameters were optimized using Adam stochastic optimisation \cite{King15} with categorical cross-entropy as the cost-function.

\section{Experiments and Results}
The data for brain MRI, breast MRI and cardiac CTA were split into 14/20, 14/20 and 6/4 training/test images, respectively. Four results were obtained for each task: one with a network trained for only that task, two with networks trained for that task and an additional task, and one with a network trained for all tasks together. Each network was trained with 25000 mini-batches per task.

No post-processing steps other than probability thresholding for evaluation purposes were performed. The results are presented on the full test set.
In brain MRI, the voxel class labels were determined by the highest class activation. The performance was evaluated per brain tissue type, using the Dice coefficient between the manual and automatic segmentations. In breast MRI and cardiac CTA, precision-recall curve analysis was performed to identify the optimal operating point, defined, for each experiment, as the highest Dice coefficient over the whole test set. The thresholds at this optimal operating point were then applied to all images.

Fig. \ref{fig:evaluation} shows the results of the described quantitative analysis, performed at intervals of 1000 mini-batches per task. As the networks learned, the obtained Dice coefficients increased and the stability of the results improved. For each segmentation task, the learning curves were similar for all experiments. Nevertheless, slight differences were visible between the obtained learning curves. To assess whether these differences were systematic or caused by the stochastic nature of CNN training, the training experiment using only brain MR data (Experiment 1) was repeated (dashed line in Fig. \ref{fig:evaluation}), showing similar inter-experiment variation.  
Fig. \ref{fig:segmentations} shows a visual comparison of results obtained for the three different tasks. For all three tasks, all four networks were able to accurately segment the target tissues. 

Confusion between tasks was very low. For the network trained with three tasks, the median percentage of voxels per scan labelled with a class alien to the target (e.g. cortical grey matter identified in breast MR) was $<0.0005\%$ for all tasks. 

\section{Discussion and Conclusions}
The results demonstrate that a single CNN architecture can be used to train CNNs able to obtain accurate results in images from different modalities, visualising different anatomy. Moreover, it is possible to train a single CNN instance that can not only segment multiple tissue classes in a single modality visualising a single anatomical structure, but also multiple classes over multiple modalities visualising multiple anatomical structures. 

In all experiments, a fixed CNN architecture with triplanar orthogonal input patches was used. We have strived to utilise recent advances in deep learning such as batch normalisation \cite{Ioff15}, Adam stochastic optimisation \cite{King15}, exponential linear units \cite{Clev16}, and very deep networks with small convolution kernels \cite{Simo14}. Furthermore, the implementation of fully connected layers as 1$\times$1 convolution layers and the omission of downsampling layers allowed fast processing of whole images compared with more time-consuming patch-based scanning \cite{Pras13,Roth15a,Wolt15,Moes16}. The ability of the CNN to adapt to different tasks suggests that small architectural changes are unlikely to have a large effect on the performance. Volumetric 3D input patches might result in increased performance, but would require a high computational load due to the increased size of the network parameter space.

The results for brain segmentation are comparable with previously published results \cite{Moes16}. Due to differences in image acquisition and patient population, the obtained results for pectoral muscle segmentation and coronary artery extraction cannot be directly compared to results reported in other studies. Nevertheless, these results appear to be in line with previously published studies \cite{Gube12,Zhen11}. No post-processing other than probability thresholding for evaluation purposes was applied. The output probabilities may be further processed, or directly used as input for further analysis, depending on the application.

Including multiple tasks in the training procedure resulted in a segmentation performance equivalent to that of a network trained specifically for the task (Fig. \ref{fig:evaluation}). Similarities between the tasks, e.g. presence of the pectoral muscle in both breast MR and cardiac CTA, or similar appearance of brain and breast tissue in T\textsubscript{1}-weighted MRI, led to very limited confusion. In future work, we will further investigate the capacity of the current architecture with more data and segmentation tasks, and investigate to what extent the representations within the CNN are shared between tasks.

\FloatBarrier

\bibliographystyle{splncs03}
\bibliography{literature}

\end{document}